\documentclass[runningheads]{llncs}
\usepackage{booktabs} 
\usepackage{xcolor} 
\usepackage{caption} 

\definecolor{LightGray}{gray}{0.95} 
\usepackage{subcaption}
\usepackage{amsmath}
\usepackage{graphicx}
\PassOptionsToPackage{hyphens}{url}
\usepackage{hyperref}
\usepackage[T1]{fontenc}
\usepackage{lmodern}
\usepackage{booktabs} 
\usepackage{xcolor} 
\usepackage{caption} 

\definecolor{headercolor}{RGB}{230,230,230} 

\begin{document}
\title{Cross-Vendor Reproducibility of Radiomics-based Machine Learning Models for Computer-aided Diagnosis}
\titlerunning{Cross-Vendor Reproducibility of Radiomics-based ML Models for CAD}
%

\author{Jatin Chaudhary*\inst{1} \and
Ivan Jambor\inst{2} \and
Hannu Aronen\inst{2} \and
Otto Ettala\inst{3} \and
Jani Saunavaara\inst{4} \and
Peter Boström\inst{3} \and
Jukka Heikkonen\inst{1} \and
Rajeev Kanth\inst{5} \and
Harri Merisaari\inst{2}
}
\authorrunning{Anonymous Authors}
\institute{Department of Computing, University of Turku, Turku, Finland 
 \and
Department of Diagnostic Radiology, University of Turku, Turku, Finland \and
Department of Urology, University of Turku, Turku, Finland \and
Department of Medical Physics, Turku University Hospital, Turku, Finland \and
Savonia University of Applied Sciences, Kuopio, Finland \\
\email{*jatin.chaudhary@utu.fi}}

%
\authorrunning{J. Chaudhary et al.}
%
%
\maketitle              

\begin{abstract}
\textbf{Background:} The reproducibility of machine-learning models in prostate cancer detection across different MRI vendors remains a significant challenge. \\
\textbf{Methods:} This study investigates Support Vector Machines (SVM) and Random Forest (RF) models trained on radiomic features extracted from T2-weighted MRI images using Pyradiomics and MRCradiomics libraries. Feature selection was performed using the maximum relevance minimum redundancy (MRMR) technique. We aimed to enhance clinical decision support through multimodal learning and feature fusion. \\
\textbf{Results:} Our SVM model, utilizing combined features from Pyradiomics and MRCradiomics, achieved an AUC of 0.74 on the Multi-Improd dataset (Siemens scanner) but decreased to 0.60 on the Philips test set. The RF model showed similar trends, with notable robustness for models using Pyradiomics features alone (AUC of 0.78 on Philips). \\
\textbf{Conclusions:} These findings demonstrate the potential of multimodal feature integration to improve the robustness and generalizability of machine-learning models for clinical decision support in prostate cancer detection. This study marks a significant step towards developing reliable AI-driven diagnostic tools that maintain efficacy across various imaging platforms.
\end{abstract}


\section{Introduction}
Thinking about the transformative era of medical diagnostics with the integration of machine learning (ML) into cancer detection opens up a new oil reserve of possibilities. The strategic application of ML in prostate cancer via Magnetic Resonance Imaging (MRI) stands at the crucial juncture of this revolution \cite{Michaely2022}, insighting towards a pivotal intersection where innovation meets urgent healthcare demands. In the field of medical imaging, radiomic feature extraction and artificial intelligence (AI) in medical imaging has experienced significant progress over the past decade, with these methods increasingly being used in MRI of prostate \cite{Bera2022}\cite{Li2022}, while reproducibility of these AI models is still not largely studied. Prostate cancer (PCa) continues to be the most common cancer among men in the western worlds and the second most common cause of death. While prostate cancer MRI is generally considered cost-effective \cite{Hao2023}, long acquisition times, high cost, and inter-center/reader variability of usual multi-parametric prostate MRI limit its wider adoption to clinical use. 

Reproducibility has recently gained more focus \cite{Keenan2022}. It is the bedrock that underpins the trustworthiness, effectiveness, and clinical acceptance of ML applications. The essence of reproducibility in this technological evolution is of paramount importance in the field of medical imaging, where number of variations hinder application and adoption of developed ML techniques to new environments.  Among others, most prominent sources of these variations are differences in MR imaging devices and MR sequence acquisition settings.  \cite{beam2020challenges} have addressed several reproducibility dimensions and clinical ramifications of AI deployment in healthcare, emphasizing reproducibility as a critical component for safeguarding patient welfare and ensuring equitable treatment. These not only cast light on the technical hurdles but also highlight the demand for improved and reliable ML methodologies in clinical practice to improve clinical practises.

Thus, the development and validation of a more cost-effective prostate MRI protocol for improved PCa risk stratification in men with elevated PSA would allow a decrease of the financial burden of implementing MRI for widespread clinical use \cite{Kim2018}. \cite{bosma2024reproducibility} have given a detailed exploration of how to consistently achieve accurate results when training AI models for medical imaging, specifically focusing on the precision in detection of PCa through MRI scans. Prostate cancer MRI has come up with reproducibility challenges, especially when ML models are implemented across different MRI ecosystems built by various vendors. Despite the advantages in terms of cost-effectiveness and patient comfort, the goal of harmonizing MRI has revealed an urgent need for uniform datasets and thorough cross-vendor validation studies to enhance model reproducibility and universality.  This quest for uniformity and validation is not merely a technical endeavor but a commitment to patient-centric care, resonating across the research community. In \cite{brembilla2022diagnostic}, it was found that the reproducibility of the model is a major concern. In examination of the variances in diagnosti imaging quality across diverse geographical and socioeconomic landscapes, \cite{renard2020variability} underscores the importance of inclusive research and validation endeavors to guarantee the widespread efficacy and applicability of ML models in varied clinical contexts. A unique bi-parametric prostate MRI, IMPROD bpMRI \cite{Ettala2022} \cite{Jambor2017}, has demonstrated potential to reduce the number of unnecessary biopsy procedures while improving detection of clinically significant prostate cancer. It provides an MR sequence, which is easily implementable and has reasonable acquisition time \cite{Jambor2015}\cite{Kahkonen2013}\cite{Ettala2022} \cite{Jambor2017}.  Using the same MR sequence in different MR devices, vendors, and imaging sites, gives possibility to study reproducibility in context where the acquisition sequence does not change, and allows potentially more reproducible and accurate results than when MR sequence is let to vary.

This article focuses on the question to confront these critical challenges by scrutinizing the reproducibility of ML models in determining the \textbf{aggressiveness of tumour} from MRI scans, spanning an array of MRI equipment from different vendors. Through a meticulous process of feature selection and the employment of a comprehensive package of evaluation metrics, this study aspires to not only affirm the attainability of reproducibility in ML models for prostate cancer diagnosis but also to facilitate their fluid incorporation into clinical workflows. In this endeavor, we evaluated two conventional machine learning models for their performance in unseen datasets using two open source radiomic feature extraction toolkits (MRCradiomics https://github.com/haanme/MRCRadiomics and pyradiomics https://github.com/AIM-Harvard/pyradiomics), to evaluate their performance when the same and a different MR device is being used. We applied our methodology in T2-weighted MRI images which are conventionally used in prostate cancer imaging, and have clinically acceptable acquisition time. This aims to accelerate the clinical acceptance of AI-powered diagnostics, ensuring that these pioneering tools improve patient care with the precision and dependability that healthcare practitioners and patients can trust.

\section{Materials and Methods}

\subsection{MRI Data Collection}
We used imaging data of men with a clinical suspicion of prostate cancer (PCa) have been enrolled in prospective, registered and completed clinical trials \\IMPROD (NCT01864135 http://mrc.utu.fi/mri/improd) and \\MULTI-IMPROD (NCT01864135 http://mrc.utu.fi/mri/multi-improd) \\PROMANEG (NCT02388126) and FLUCIPRO (NCT02002455), and prostate cancer datasets used earlier in \cite{Jambor2015} and \cite{Kahkonen2013}. The trials were approved by the Institutional Review Board (IRB) and each enrolled man gave written informed consent. In total, 637 men with a clinical suspicion of PCa were included in the pooled study sample. The entire imaging protocol included both Diffusion Weighted Imaging with IMPROD protocol \cite{Ettala2022} \cite{Jambor2017} with shimming and calibration, took 15–17 minutes per patient. While the IMPROD bpMRI protocol consisted of optimized T2-weighted (axial and sagittal) and three separate diffusion weighted imaging (DWI) acquisitions, only axial T2-weighted images were used in this study, to evaluate performance in situation where only a part would be obtained for sake of faster MR examination. All the datasets were scanned with same MR device (Siemens MAGNETOM Verio 3T), while portion of one dataset was scanned with MRI device from different vendor (Philips Ingenia 3T).

\subsection{Image Data Post-processing}

Prostate cancer aggressiveness was graded using Gleason Grade Groups (GGG) \cite{Loeb2016} based on samples from either with systemic or targeted biopsy. Radiomic feature extraction was then performed using pyradiomics package \cite{VanGriethuysen2017}, and MRCradiomics package for repeatable radiomics \cite{Merisaari2020}. We extracted total of 2693 radiomic feature values (11075 with pyradiomics package and 1618 with MRCradiomics package) from lesions in each subject. We splitted the data to training, validation and unseen test datasets, with proportions of 54.6\%, 13.6\%, and 31.8\%, correspondingly. 

\subsection{Feature Selection}
Feature selection using the Maximum Relevance Minimum Redundancy (MRMR) has been done to avoid redundancy and select features that provide the most insight into the predictive power of features to the model and avoiding overlap or redundancy of the features\cite{peng2005}. The first set included the top 40 features that were automatically selected by MRMR and later exposed to univariate analysis for the assessment of individual variable performance on the training dataset and validation datasets. This led the distillation of the feature set down to 14 most predictive variables through this rigorous process. Thus, these parameters were individually significant and presented a synergistic effect in increasing further model precision to ensure the possibility of model replicability on different MRI platforms. The final selection of these features under our commitment to a purely data-driven approach in model training, ensuring that the selected features are actually effective in expressing, with robustness and reliability, the biological characteristics linked to a diagnosis of prostate cancer\cite{ding2005}\cite{peng2005}. MRMR algorithm from scikit-learn package (version 0.2.8) was used to select top 14 features to be used in the subsequent analysis, where number of selected features was determined by individual performance in training and validation datasets. We applied feature selection for packages of MRCradiomics and pyradiomics individually, and jointly to create three sets of selected radiomic features for our evaluations.

\subsection{Predictive ML Models}
In our experimentation, we divided the cohort of 637 participants into three separate groups: 434 were used for training and internal checks, following an 80-20 split for effective learning and thorough self-assessment. The rest were set aside for external tests, with 180 having the same scanner vendor (Siemens) and 23 being scanned with scanner from different vendor (Philips), to rigorously test how well our trained model works across different scenarios. 

We evaluated two conventionally used ML techniques, building a Support Vector Machine (SVM) model and a Random Forest model. For the former, given that SVM is known for its robust binary classification\cite{cortes1995support}, we imputed the missing data and adjusted values carefully using median imputation, which is crucial for the SVM to work well with lots of different features\cite{hsu2003practical}. For the latter, Random Forest was employed. The same data splits and feature selection protocol was used to train both of the techniques. 

\section{Result}
The selected features for Pyradiomics, MRCradiomics, and their combination (referred to as Pyradiomics+MRCradiomics) are detailed in Tables 1, 2, and 3 (in appendix), respectively.

When applying the SVM model to the combined features from Pyradiomics and MRCradiomics, an AUC score of 0.74 (90\% CI: [0.68, 0.79]) was achieved on the Multi-Improd dataset using the same scanner. In contrast, the performance on the Philips test set yielded a lower AUC of 0.35 (90\% CI: [0.22, 0.49]), indicating the influence of different scanners on model accuracy. The Random Forest model exhibited similar results, with an AUC of 0.73 (90\% CI: [0.68, 0.79]) on the Multi-Improd set and an AUC of 0.60 (90\% CI: [0.44, 0.76]) on the Philips test set.
\begin{figure}[h]
\centering
\includegraphics[width=0.4\linewidth]{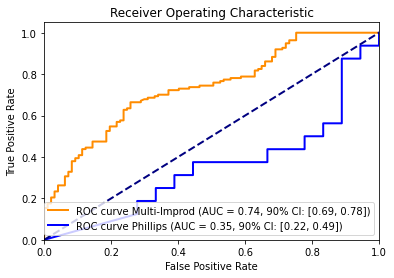}
\caption{Result of SVM trained over Pyradiomics and MRCradiomics}
\label{fig:svm}
\end{figure}

\begin{figure}[h]
\centering
\includegraphics[width=0.4\linewidth]{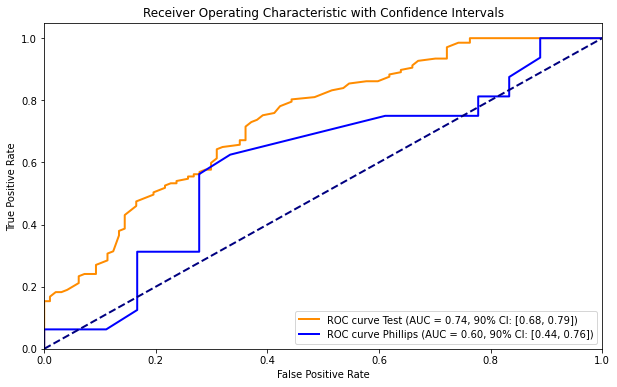}
\caption{Results of Random Forest trained over Pyradiomics and MRCradiomics}
\label{fig:randomforest}
\end{figure}

Further analysis of models trained exclusively on Pyradiomics or MRCradiomics features for reproducibility on Philips data revealed notable differences. For Pyradiomics features alone, the Random Forest model achieved a robust AUC of 0.68 (90\% CI: [0.61, 0.74]) on the Multi-Improd set and AUC of 0.78 (90\% CI: [0.61, 0.94]) on the Philips set, highlighting the significant predictive capability of Pyradiomics-derived features. The SVM model trained on Pyradiomics features alone yielded an AUC of 0.77 (90\% CI: [0.59, 0.92]) on the Philips test set but performed poorly on the Multi-Improd test set, with an AUC of 0.60 (90\% CI: [0.53, 0.66]).
\begin{figure}
\centering
\includegraphics[width=0.4\linewidth]{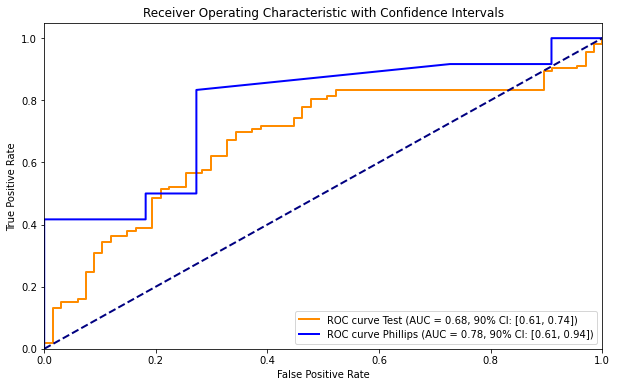}
\caption{Result of Random Forest trained over Pyradiomics}
\label{fig:sub1}
\end{figure}

\begin{figure}
\centering
\includegraphics[width=0.4\linewidth]{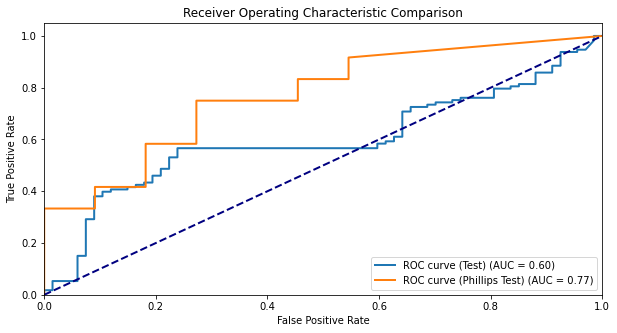}
\caption{Results of SVM trained over Pyradiomics}
\label{fig:sub2}
\end{figure}

Conversely, utilizing only MRCradiomics features resulted in varied outcomes across models. The Random Forest model achieved an AUC of 0.56 (90\% CI: [0.40, 0.73]) on the Philips set and an AUC of 0.68 (90\% CI: [0.63, 0.74]) on the Multi-Improd test set. The SVM model demonstrated an AUC of 0.53 (90\% CI: [0.36, 0.71]) on the Philips set and an AUC of 0.73 (90\% CI: [0.67, 0.78]) on the Multi-Improd test set.
 Figure 1
\begin{figure}[h]
\centering
\includegraphics[width=0.4\linewidth]{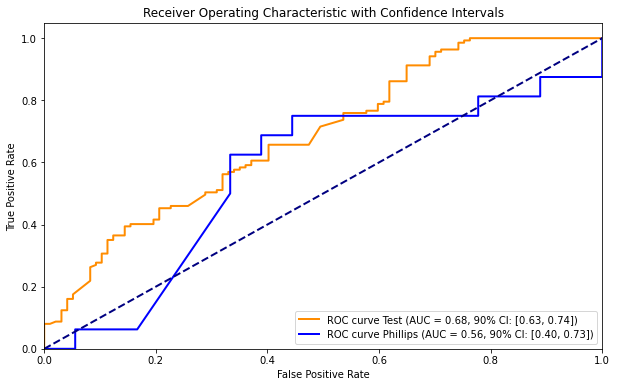}
\caption{Result of Random Forest trained over MRCradiomics.}
\label{fig:fig5}
\end{figure}

\begin{figure}[h]
\centering
\includegraphics[width=0.4\linewidth]{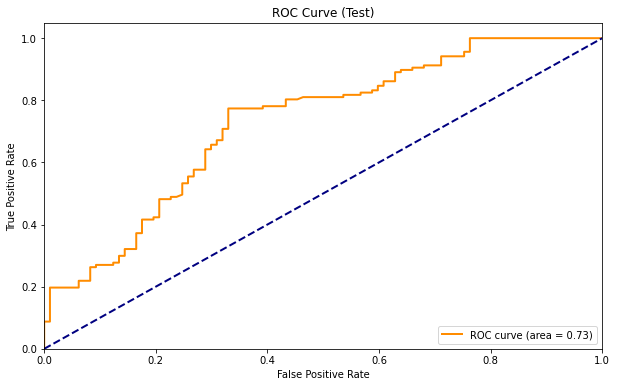}
\caption{Results of SVM trained over MRCradiomics and tested on Multi-Improd Dataset.}
\label{fig:fig6}
\end{figure}

\begin{figure}[h]
\centering
\includegraphics[width=0.4\linewidth]{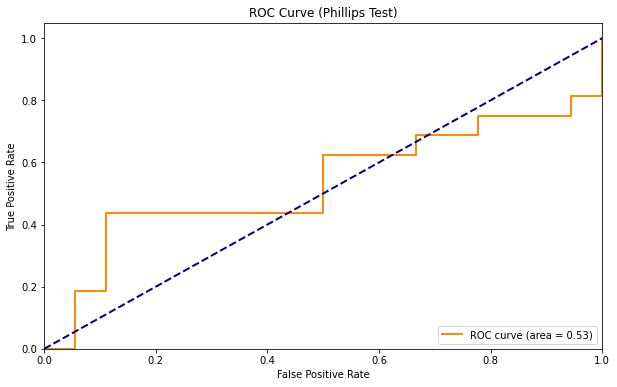}
\caption{Results of SVM trained over MRCradiomics and tested on Philips Dataset.}
\label{fig:fig7}
\end{figure}

\section{Discussion}
In this comprehensive study, we undertook a detailed exploration of machine learning models for the prediction of aggressiveness in prostate cancer, leveraging advanced radiomics features derived from pyradiomics and MRC radiomics packages\cite{Qi2023} \cite{Gresser2022}\cite{Wang2022}. The investigation encompassed an extensive range of hyper-parameters for RandomForest and SVM algorithms to optimize performance, guided by methodical tuning processes including RandomizedSearchCV for RandomForest and GridSearchCV for SVM\cite{Qi2023} \cite{Liu2020}. These processes targeted the ROC AUC metric for evaluation, underpinning our commitment to precision in model efficacy assessment\cite{Qi2023} \cite{Gresser2022}. Our analysis demonstrate the efficacy of combining features from both pyradiomics and MRCradiomics, unveiling a promising potential to enhance predictive performance in machine learning applications within the realm of healthcare diagnostics\cite{Qi2023} \cite{Liu2020} \cite{Gresser2022}.

The SVM model, which is a supervised learning algorithm primarily used for classification tasks, operates by finding the hyperplane that best separates different classes in the feature space \cite{huang2002support}. It relies on kernel functions to handle non-linear boundaries, thereby making it quite effective in high-dimensional spaces \cite{wang2009empirical}. The SVM model achieved a relatively high AUC score of 0.74 on the Multi-Improd set when utilizing combined features from Pyradiomics and MRCradiomics, demonstrating its ability to leverage a diverse feature set effectively \cite{niaf2014kernel}. However, its performance dropped significantly on the Philips test set (AUC 0.60), indicating a susceptibility to variations introduced by different scanners \cite{candelieri2014hyper}. In contrast, the Random Forest model, an ensemble learning method based on the construction of multiple decision trees, showed a similar trend with an AUC of 0.73 on the Multi-Improd set and 0.60 on the Philips test set \cite{madzarov2009multi}. The key advantage of Random Forest lies in its aggregation of predictions from various decision trees, which enhances its robustness and reduces the risk of overfitting \cite{gunakala2023comparative}. This ensemble approach also allows for better generalization across datasets with varying characteristics, although it still showed sensitivity to scanner variations similar to the SVM model \cite{gurram2013sparse}. The Random Forest model's superior performance on the Philips set with Pyradiomics features (AUC 0.78) can be attributed to its ability to identify and prioritize the most predictive features, thereby offering insights into the specific characteristics that are most relevant for distinguishing between classes \cite{jiang2019ssem}. The SVM model, despite its high performance in certain scenarios, lacks this level of transparency \cite{wang2009empirical}. The decision boundaries created by SVMs are often difficult to interpret, especially when complex kernel functions are employed \cite{haasdonk2005feature}. While SVMs can effectively handle non-linear relationships within the data, the trade-off is a loss of explainability, which can be a drawback when understanding the model's decision-making process is crucial \cite{huang2002support}.

In our study focused on Clinical Decision Support (CDS) for cancer detection, the impact of radiomic feature sets—Pyradiomics, MRCradiomics, and their combination—on the model's performance and reproducibility was critically examined. The integration of features from both Pyradiomics and MRCradiomics demonstrated that while combining these features yielded competitive AUC scores on the Multi-Improd set, it highlighted significant variability when tested across different scanners. Specifically, the AUC scores dropped notably on the Philips test set, underlining the challenge of scanner-induced variability in the reproducibility of radiomic features. This discrepancy emphasizes the necessity for rigorous validation of radiomic features across diverse imaging platforms to ensure robust and reliable clinical decision support systems. Furthermore, the analysis of models trained solely on Pyradiomics or MRCradiomics features revealed distinct patterns in predictive performance. Pyradiomics-derived features consistently showed higher predictive power and robustness, as evidenced by higher AUC scores on the Philips test set, compared to MRCradiomics features. This suggests that Pyradiomics features might capture more reproducible and clinically relevant information, which is crucial for the generalizability of CDS systems. The explainability of our results is thus anchored in the reproducibility of these feature sets; Pyradiomics features, in particular, offer a more stable basis for CDS in cancer detection, enhancing the reliability and accuracy of clinical decisions across different imaging conditions.
The observed differences in model performance between the Multi-Improd and Philips test sets can be attributed to several critical factors, primarily revolving around the imaging scanners used and the dataset sizes. The Multi-Improd dataset, captured using a Siemens machine, comprised 180 cases, providing a robust and diverse sample set for testing. In contrast, the Philips test set contained only 23 cases, significantly limiting the variability and generalizability of the models. The discrepancy in sample size likely contributed to the higher variance and lower AUC scores observed in the Philips test set, underscoring the challenges of training machine learning models on smaller datasets. The difference in imaging technology between Siemens and Philips scanners could introduce variations in image resolution, contrast, and other technical parameters, which may affect the radiomic features extracted and, consequently, the model's predictive performance.

\section{Conclusion}
In this study, we have comprehensively examined the efficacy of machine learning models in predicting the aggressiveness of prostate cancer through the utilization of advanced radiomics features from Pyradiomics and MRCradiomics packages. By exploring a broad spectrum of hyper-parameters for Random Forest and SVM algorithms, we sought to optimize model performance, meticulously tuning them via RandomizedSearchCV and GridSearchCV to prioritize ROC AUC metrics. Our findings underscore the potential of combining features from both Pyradiomics and MRCradiomics to enhance predictive performance in healthcare diagnostics, particularly in the realm of prostate cancer detection.

The evaluation of our models highlighted significant variability in performance across different imaging platforms, with the Multi-Improd dataset (Siemens scanner) showing higher AUC scores compared to the Philips test set. This discrepancy can be attributed to differences in dataset sizes and the inherent variations in imaging technology between Siemens and Philips scanners. Specifically, the larger and more diverse Multi-Improd dataset provided a more robust training ground for our models, while the smaller Philips dataset introduced higher variance and lower generalizability. These results emphasize the critical need for rigorous validation of radiomic features across various imaging platforms to ensure the reliability and applicability of Clinical Decision Support systems. By demonstrating the superior predictive power of Pyradiomics-derived features, our study highlights their potential as a stable and highly reproducible features.
\bibliographystyle{splncs04}
\bibliography{ref.bib}

\section*{Appendix}

\begin{table}[h]
\centering
\begin{tabular}{p{0.7\textwidth} | p{0.1\textwidth}}
\hline
Feature Name & Value \\
\hline

log\_sigma\_1\_GLDM\_LargeDependence & 0.672 \\
log\_sigma\_2\_GLRLM\_ShortRunHighLevelEmph & 0.651 \\
log\_sigma\_2\_GLRLM\_HighGrayLevelRunEmph & 0.649 \\
log\_sigma\_2\_GLDM\_HighGrayLevelEmph & 0.649 \\
log\_sigma\_1\_GLSZM\_SizeZoneNonUniformity & 0.647 \\
log\_sigma\_2\_GLSZM\_HighGrayLevelZoneEmph & 0.646 \\
log\_sigma\_1\_GLRLM\_ShortRunHighLevelEmph & 0.646 \\
wavelet\_HLL\_GLDM\_LargeDependence & 0.645 \\
log\_sigma\_1\_GLRLM\_HighGrayLevelRunEmph & 0.645 \\
log\_sigma\_1\_GLDM\_HighGrayLevelEmph & 0.644 \\
log\_sigma\_1\_GLCM\_Autocorrelation & 0.644 \\
diagnostics\_Mask\_interpolated\_Maximum & 0.644 \\
original\_firstorder\_Maximum & 0.644 \\
log\_sigma\_1\_firstorder\_Range & 0.642 \\
original\_firstorder\_Range & 0.641 \\
\hline
\end{tabular}
\caption{Selected radiomic features extracted with pyradiomics package. Features were selected with MRMR method using T2-weighted images of 434 prostate cancer subjects, their respective descriptions, package names, and AUC performance in validation set. }
\label{table:feature_values}
\end{table}

\begin{table}[h]
\centering
\begin{tabular}{p{0.7\textwidth} | p{0.1\textwidth}}
\hline
Feature Name  & AUC \\
\hline
0.05\_No\_corners\_ROI & 0.679 \\
0.01\_No\_corners\_ROI & 0.679 \\
0.05\_No\_corners\_ROI & 0.678 \\
0.01\_No\_corners\_ROI & 0.677 \\
0.01\_No\_corners\_ROI & 0.674 \\
objprops\_N\_objs & 0.673 \\
objprops\_Per\_IQR\_mm & 0.671 \\
0.05\_No\_corners\_ROI & 0.670 \\
objprops\_Int\_IQR & 0.668 \\
Frangi\_objprops\_Int\_SD & 0.667 \\
Frangi\_objprops\_Int\_IQR & 0.665 \\
Scharr\_objprops\_Area\_median\_mm2 & 0.665 \\
Range & 0.663 \\
Scharr\_objprops\_Area\_mean\_mm2 & 0.663 \\
Hessian\_0.025\_15.0\_objprops\_Int\_SD & 0.658 \\
\hline
\end{tabular}
\caption{Selected radiomic features extracted with MRCradiomics package. Features were selected with MRMR method using T2-weighted images of 434 prostate cancer subjects, their respective descriptions, package names, and AUC performance in validation set.}
\label{table:description_auc}
\end{table}

\begin{table}[h]
\centering
\begin{tabular}{p{0.6\textwidth} | p{0.1\textwidth} | p{0.1\textwidth}}
\hline
Description & Package & AUC \\
\hline
Wavelet LLL gldm DependenceEntropy & PYR & 0.697 \\
Wavelet LLL glszm ZoneEntropy & PYR & 0.696 \\
Wavelet LLL glcm JointEntropy & PYR & 0.694 \\
Original shape LeastAxisLength & PYR & 0.690 \\
Log sigma 2.0 mm 3D gldm DependenceEntropy & PYR & 0.686 \\
1 mm original gldm DependenceEntropy & PYR & 0.682 \\
1 mm Wavelet LHL glcm Idmn & PYR & 0.681 \\
1 mm log sigma 3.0 mm 3D glcm DifferenceEntropy & PYR & 0.678 \\
1 mm Wavelet LHH gldm DependenceVariance & PYR & 0.677 \\
1 mm Wavelet HHL glcm Imc1 & PYR & 0.677 \\
1 mm Wavelet LHL glcm Idn & PYR & 0.676 \\
Harris-Stephens corner-edge b4 ks7 k0.50 Corner density primary & MRCR & 0.666 \\
Harris-Stephens corner-edge b4 ks7 k0.50 Corner density mean & MRCR & 0.662 \\
Scharr filtered object properties: Number of objects & MRCR & 0.661 \\
\hline
\end{tabular}
\caption{Selected radiomic features extracted with pyradiomics(PYR) and MRCradiomics(MRCR) packages. Features were selected with MRMR method using T2-weighted images of 434 prostate cancer subjects, their respective descriptions, package names, and AUC performance in validation set. (PY: Pyradiomics and MRC: MRCradiomics)}
\label{table:description_package_auc}
\end{table}

\end{document}